\documentclass[11pt]{article}
\usepackage[margin=1in]{geometry}
\usepackage{amsmath,amssymb,amsfonts,bm}
\usepackage{booktabs,caption,subcaption}
\usepackage{xcolor}
\usepackage{colortbl}
\usepackage{multirow}
\usepackage{adjustbox}
\usepackage{textcomp}
\usepackage{siunitx}
\usepackage{algorithm}
\usepackage{algpseudocode}
\usepackage{natbib}
\usepackage{hyperref}
\usepackage{authblk}
\usepackage{amsmath}
\hypersetup{colorlinks=true, citecolor=blue, linkcolor=blue, urlcolor=blue}

\title{A General-Purpose Dual-Path Interaction Framework for Multi-View Medical Image Segmentation: Alleviating Data Volume Dependency in Small-Sample Regimes}

\author{
Shuhuan Wang$^{\text{a},1}$, Yuanyuan Cui$^{\text{b},1}$, Pengju Wang$^{\text{a}}$, Huimin Guo$^{\text{a}}$, Ling Yue$^{\text{b}}$, Yin Gu$^{\text{a}}$, \\
Yanchao Zhang$^{\text{b}}$, Wei Feng$^{\text{b}}$, Yunhe Liu$^{\text{b}}$, Wei Qian$^{\text{b}}$, Mingjun Xu$^{\text{d},*}$, Shuang Liu$^{\text{b},*}$, He Ma$^{\text{a},c,*}$ 
}
\affil{
$^{\text{a}}$ College of Medicine and Biological Information Engineering, Northeastern University, Shenyang, Liaoning, 110169, China \\
$^{\text{b}}$ Department of Ultrasound, The Fourth Affiliated Hospital of China Medical University, Shenyang, Liaoning, 110000, China \\
$^{\text{c}}$ The Key Laboratory of Intelligent Computing in Medical Image, Ministry of Education, Northeastern University, Shenyang, Liaoning, 110819, China \\
$^{\text{d}}$ Department of Cardiology, Qilu Hospital of Shandong University \\
$^{\text{1}}$ These authors contributed equally
}

\begin{document}
\maketitle
\begin{abstract}
The performance of medical image segmentation depends on a large amount of annotated data, and most segmentation models experience a significant drop in performance in low-sample-size scenarios. Existing semi-supervised and weakly supervised methods rely on specific network architectures, while traditional multi-view-assisted segmentation methods require strictly paired samples, which limits their clinical applications. To address these issues, we propose DPIF, a general-purpose plug-and-play dual-path interaction framework compatible with arbitrary encoder-decoder segmentation networks, which leverages cross-view anatomical complementarity to reduce annotation reliance. DPIF first performs Sinkhorn optimal transport-based cross-view distribution alignment to establish global semantic correspondence without sample-level pairing. It then adopts conditional mutual information-guided feature decoupling to enable bidirectional knowledge transfer only within the shared semantic space, suppressing view-specific noise. Finally, persistent homology-based topological regularization encodes anatomical invariants as unsupervised priors to mitigate overfitting. Experiments were conducted using 24 baseline models on a private multicenter echocardiography dataset and two public datasets. The results show that DPIF-enhanced models outperform single-view baseline models when trained on comparable amounts of data. Notably, most dual-path variants trained on 40\% - 60\% of the data achieve performance comparable to that of single-view models trained on the full dataset. This demonstrates that DPIF provides a general-purpose and scalable solution for data-scarce multi-view medical image segmentation problems.
\end{abstract}

\section{Introduction}
\label{sec1}
Accurate medical image segmentation is the foundation for clinical quantitative analysis, disease diagnosis, and surgical planning. Although deep learning has advanced this field, its performance still depends on large-scale datasets with pixel-level annotations\cite{1}. Unlike natural images, annotating medical images requires specialized clinical knowledge, making the process both time-consuming and costly. For multi-view imaging protocols, each view must be annotated separately, leading to a significant increase in workload and costs. Therefore, the scarcity of annotated data remains a major bottleneck for clinical translation, particularly for rare diseases\cite{2}.

In standard clinical practice, images of the same anatomical structure are typically acquired from multiple angles to enable a comprehensive assessment of lesions\cite{3}. For example, a standard echocardiogram simultaneously acquires both the long-axis and short-axis views of the left ventricle. Although these views contain highly complementary anatomical information, they differ significantly in terms of geometric structure and imaging characteristics\cite{4}. Current mainstream segmentation methods still process each view independently, failing to fully leverage the inherent structural and semantic relationships between views. Most existing multi-view fusion methods employ simple strategies\cite{5,6}. These methods overlook the substantial spatial structural differences and variations in semantic distribution across views, often introducing redundant noise and yielding only limited performance improvements. More critically, most multi-view frameworks are tightly coupled with specific backbone architectures, which hinders their rapid adaptation to new tasks and poses challenges for cross-scenario deployment\cite{7}.

To address these fundamental limitations, we propose a general-purpose dual-path interaction framework. Its core idea is to leverage the complementarity of cross-view anatomical structures as an alternative to expanding the annotated dataset. The framework features a universal, plug-and-play design that is compatible with various segmentation backbone models and achieves high-precision segmentation with only a small number of annotated samples. The main contributions of this study are summarized as follows:

We propose a general-purpose dual-path interaction framework designed to reduce reliance on segmented annotated data. This framework can be integrated into any encoder-decoder Segmentation network.
We propose a cross-view alignment mechanism based on optimal transport theory. It establishes semantic correspondences by minimizing the Wasserstein distance, thereby enabling distribution-level alignment of unpaired data.
We propose a two-layer decoupling strategy for cross-view knowledge transfer. Under conditional mutual information constraints, features are decomposed into shared and view-specific components, and knowledge transfer and enhancement are performed only in the shared space to suppress noise propagation.
We propose a regularization method based on cross-view topological consistency. This method uses anatomical topological invariants as shared prior information to impose constraints, thereby narrowing the hypothesis space and mitigating the problem of overfitting in small-sample settings.

\section{Related WORK}

\subsection{Multi-view Medical Image Segmentation}

Existing multi-view medical image segmentation methods can be broadly categorized into three paradigms: input-level, feature-level, and decision-level fusion. Input-level fusion stacks multi-view images as multi-channel inputs but ignores spatial differences between views. Feature-level fusion directly concatenates or superimposes intermediate features within the network, but carries the risk of introducing view-specific noise and artifacts\cite{8}. Decision-level fusion, while capable of merging independent prediction outputs, discards complementary semantic information present in the hidden layers\cite{9}. Although improvements based on attention mechanisms have shown some promise, most fusion schemes remain constrained by their network architectures, making it difficult to transfer them across different segmentation pipelines\cite{10}. Furthermore, most of these schemes are designed for fully annotated datasets and perform poorly when data is limited. Crucially, nearly all existing methods require strictly paired multi-view patient data, which imposes severe constraints on real-world clinical inputs.

\subsection{Small-Sample Medical Image Segmentation}

To address the issue of scarce annotations, existing research primarily follows three approaches: semi-supervised learning, weakly supervised learning, and few-shot learning. Semi-supervised methods utilize large amounts of unlabeled data to aid training, but still rely on a certain scale of unlabeled data, thus imposing requirements on the data volume\cite{11}. Weakly supervised methods employ weak labels, such as point annotations and bounding box annotations, to reduce annotation costs. However, due to the limitations imposed by the simplicity of these labels, there is an upper bound on accuracy.\cite{12} Few-shot learning methods aim to achieve generalization with small datasets through cross-domain knowledge transfer. However, due to significant variations across medical imaging domains, their practical effectiveness remains uncertain\cite{2}. It is worth noting that existing methods rarely exploit the inherent correlations among multi-view images of the same object, nor have they established a universal interaction scheme that can be reused across models. Meanwhile, the cost of acquiring multi-view images in clinical settings is far lower than the cost of annotating data required for a single view, and this potential has yet to be fully utilized.

\section{Method}

\begin{figure*}[htbp] 
\vspace{-0.3cm}
\centering  
\centerline{\includegraphics[width=\linewidth]{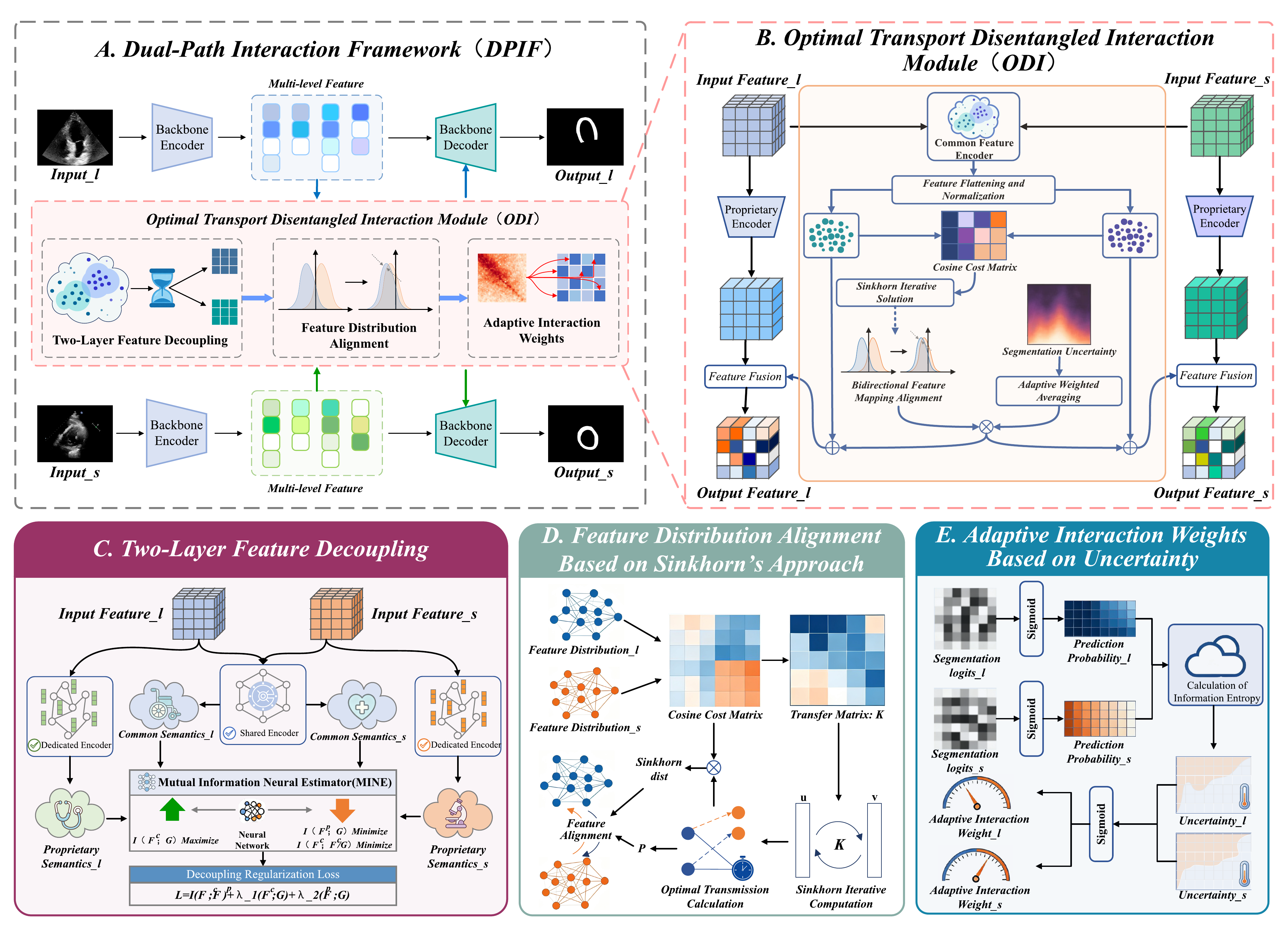}}
\caption{DPIF Framework and Functional Modules. (A) Schematic diagram of the dual-branch interaction framework deployment. (B) Core interaction module, used to connect the two segmentation pipelines. (C) Two-layer feature decoupling method, which distinguishes between shared and proprietary features. (D) Feature distribution alignment algorithm, which enables bidirectional semantic interaction and enhancement. (E) Segmentation uncertainty-guided weight interaction method, which enables adaptive adjustment of interaction intensity.}
\vspace{-0.5cm}
\label{fig1}  
\end{figure*}

\subsection{Overview}

As shown in Fig \ref{fig1}, our proposed Dual-Path Interaction Framework (DPIF) adopts a symmetric dual-branch architecture, in which the backbone networks share the same structure but are initialized independently. Each branch receives a single-view image for segmenting the target anatomical structure. The training process is divided into three stages. In the feature encoding stage, a dual-branch encoder extracts multiscale features. Subsequently, the ODI module is invoked at designated encoder-decoder layers to perform feature decoupling, unpaired distribution alignment, and adaptive bidirectional feature enhancement, after which the enhanced features are fed back to the decoder. Meanwhile, a topological regularization loss serves as an anatomical prior to ensure topological consistency across different views. In the final decoding and joint training stage, each decoder upsamples and aggregates the features to generate segmentation outputs and soft probability maps. The entire framework is optimized via a composite loss function. During the inference stage, the interaction module is disabled, and the two branches operate independently.

\subsection{Optimal Transmission Decoupling Interaction Module}

The ODI module is the core functional unit of DPIF, enabling cross-view feature interaction. The specific workflow of this module consists of three consecutive steps: two-layer feature decoupling, unpaired distribution alignment, and adaptive bidirectional feature enhancement.

\subsubsection{Two-Layer Feature Decoupling}

Multi-view medical images contain both shared anatomical semantics and view-specific noise, intensity shifts, and artifacts. Directly propagating raw features across branches reduces the quality of feature interaction, especially when data is limited. To address this issue, our proposed two-layer feature decoupling method decomposes the input features of each view into two mutually exclusive components: a shared semantic component for cross-view information exchange, and a view-specific component that is excluded from inter-branch interaction. Each branch employs two parallel \(1\times1\) convolutional encoders. One encoder, with shared weights, extracts shared semantic features \(F^c\) to construct a unified semantic space, while the view-independent encoder generates view-specific features \(F^p\):

\begin{equation}
F^c = \text{Enc}_c(F),\quad F^p = \text{Enc}_p(F)
\label{eq1}
\end{equation}

To guarantee effective decoupling, we introduce a conditional mutual information constraint as a regularization term:

\begin{equation}
\mathcal{L}_{\text{decouple}} = I(F^c;F^p|G) - \lambda_1 I(F^c;G) - \lambda_2 I(F^p;G)
\label{eq2}
\end{equation}

where \(I(\cdot;\cdot)\) denotes mutual information and G is the segmentation label. This loss maximizes the dependence between shared components and segmentation labels, and minimizes conditional mutual‑information between \(F^c\) and \(F^p\) given labels, guaranteeing their statistical independence and preventing noise from polluting the shared semantic space.

\subsubsection{Sinkhorn-Based Unpaired Distribution Alignment}

Given the scarcity of multi-view data for sample-level pairing, which limits the scope of application, we reformulate cross-view semantic matching as an sinkhorn-Based optimal transport problem to achieve global alignment at the distribution level\cite{13}. First, the shared semantic features from both branches are flattened into sets of feature points. For the long-axis branch, the feature set is \(X = \{x_i\}_{i=1}^{N_l}\) with \(x_i \in \mathbb{R}^{C_c}\), corresponding to the discrete distribution \(\mu = \frac{1}{N_l}\sum_{i=1}^{N_l}\delta_{x_i}\). For the short-axis branch, the feature set is \(Y = \{y_j\}_{j=1}^{N_s}\) with \(y_j \in \mathbb{R}^{C_c}\), corresponding to the discrete distribution \(\nu = \frac{1}{N_s}\sum_{j=1}^{N_s}\delta_{y_j}\). Here, \(N_l = B_l \cdot H_l \cdot W_l\) and \(N_s = B_s \cdot H_s \cdot W_s\) denote the total number of feature points in the two branches, respectively. The two values can differ, making the formulation compatible with unpaired settings.
We define the cost matrix \(C \in \mathbb{R}^{N_l \times N_s}\), where each element \(C_{ij} = \|x_i - y_j\|_2^2\) represents the squared Euclidean distance between two feature points, i.e., the transport cost. The optimal transport problem aims to find a transport plan \(P \in \Pi(\mu,\nu)\) that satisfies the marginal distribution constraints while minimizing the total transport cost:
\(\min_{P \in \Pi(\mu,\nu)} \langle P, C \rangle_F\).
To improve computational efficiency and ensure differentiability, we introduce an entropy regularization term, converting the problem into a Sinkhorn distance optimization:
\begin{equation}
d_{\text{Sinkhorn}}(\mu,\nu) = \min_{P \in \Pi(\mu,\nu)} \langle P, C \rangle_F - \varepsilon H(P)
\label{eq3}
\end{equation}
where \(H(P) = -\sum_{i,j} P_{ij}\log P_{ij}\) is the entropy of the transport plan and \(\varepsilon\) is the regularization coefficient. This problem can be efficiently solved via the Sinkhorn–Knopp iterative algorithm to yield the optimal transport plan \(P^*\).
Based on the optimal transport plan, we achieve bidirectional feature mapping and enhancement. Short-axis features are mapped to the long-axis feature space to generate complementary features for the long-axis view: \(F_{l\leftarrow s}^c = P^* \cdot F_s^c\). Similarly, complementary features for the short-axis view are obtained as \(F_{s\leftarrow l}^c = (P^*)^T \cdot F_l^c\). This process establishes semantic correspondence between the two views at the global distribution level, enabling knowledge transfer without requiring sample-level pairing.

\subsubsection{Adaptive Interaction Weights Based on Segmentation Uncertainty}

In clinical practice, there are inherent differences in imaging quality and segmentation difficulty across different views. Fixed, symmetric interaction cannot adapt to such differences, often resulting in underutilization of high-quality information while allowing low-quality information to cause excessive interference. To address this issue, we designed an energy function based on segmentation uncertainty to dynamically adjust the intensity of bidirectional interaction, thereby enabling the directed transmission of high-confidence features to the low-confidence branch.

First, we calculate the segmentation uncertainty for each view based on its predicted map:

\begin{equation}
E = -\frac{1}{N}\sum_{i=1}^N \left( p_i\log p_i + (1-p_i)\log(1-p_i) \right)
\label{eq4}
\end{equation}

Here, \(p_i\) is the predicted probability that pixel i belongs to the target region, and \(E \in [0,1]\). A higher value indicates greater segmentation uncertainty and lower confidence.

Based on the difference in uncertainty between the two branches, adaptive interaction weights are generated using the sigmoid function:

\begin{equation}
\alpha_l = \alpha_0 \cdot \sigma\left( \frac{E_l - E_s}{\tau} \right),\quad \alpha_s = \alpha_0 \cdot \sigma\left( \frac{E_s - E_l}{\tau} \right)
\label{eq5}
\end{equation}

where \(\alpha_0\) is the base interaction strength, and \(\tau\) is the temperature coefficient that controls the smoothness of weight adjustments. When the confidence in the long-axis segmentation is significantly lower than that of the short-axis segmentation, \(\alpha_l\) increases, causing the long-axis branch to receive more guidance from the short-axis branch, the opposite is also true.

\subsection{Cross-View Topological Consistency Regularization Loss}

In small-sample scenarios, limited training coverage may induce spurious pixel-level correlations and produce unreasonable segmentation results, while pixel-based loss functions struggle to mitigate this phenomenon. We consider that anatomical structures possess inherent topological invariance, and that key topological statistics across multi-view projections exhibit high consistency, which can serve as unsupervised priors to narrow the model’s hypothesis space and alleviate overfitting. Therefore, drawing inspiration from persistent homology-based topological data analysis methods, we propose a cross-view topological consistency regularization loss\cite{14}.
Starting from the segmentation probability map of each view, we obtain a persistent diagram (PD) through sub-level set filtering, where the points encode topological features characterized by thresholds and corresponding persistent values. Higher persistence values indicate more stable and anatomically plausible structures. Since the exact persistence homology calculation is non-differentiable, we employ a differentiable approximation method: applying multiscale Gaussian smoothing to the probability graph to obtain an approximate persistence graph. The second-order Wasserstein distance between the persistence graphs of the two views forms our topological loss function:

\begin{equation}
\mathcal{L}_{\text{topo}} = W_2(\text{PD}(P_l), \text{PD}(P_s))
\label{eq6}
\end{equation}
where \(\text{PD}(\cdot)\) denotes the persistent diagram from which the segmentation results are extracted, and \(W_2(\cdot)\) is the second-order Wasserstein distance. This loss ensures that the segmentation results from the two viewpoints remain consistent in their topological statistical distributions, guiding the model to learn segmentation patterns that conform to anatomical priors.

\subsection{Joint Loss Function}

The total loss of DPIF integrates four terms to balance segmentation performance, feature‑decoupling quality, topological consistency and cross‑view knowledge complementarity for end‑to‑end joint optimization:
\begin{equation}
\mathcal{L}{\text{total}} = \mathcal{L}{\text{seg}} + \lambda_d \mathcal{L}{\text{decouple}} + \lambda_t \mathcal{L}{\text{topo}} + \lambda_k \mathcal{L}_{\text{kd}}
\label{eq7}
\end{equation}
where \(\lambda_d\), \(\lambda_t\), \(\lambda_k\) denote loss‑weight hyper‑parameters. The primary segmentation loss \(\mathcal{L}_{\text{seg}}\) combines Dice loss and binary cross‑entropy (BCE) to handle medical class imbalance, computed and averaged independently for each view.The decoupling loss enforces mutual‑information constraints: it minimizes conditional mutual information between shared/view‑specific features given labels, maximizes relevance between shared features and annotations, and suppresses task‑irrelevant view‑specific signals:
\begin{equation}
\mathcal{L}_{\text{decouple}} = I(F^c; F^p \mid G) - \lambda_1 I(F^c; G) - \lambda_2 I(F^p; G)
\label{eq8}
\end{equation}
Here \(I(\cdot;\cdot)\) and \(I(\cdot;\cdot\mid\cdot)\) represent mutual and conditional mutual information. \(F^c\), \(F^p\) are shared and view‑specific features, G is the segmentation label, and \(\lambda_1,\lambda_2\) are weighting coefficients.
The topological consistency loss regularizes cross‑view topological statistics via persistent‑homology priors to shrink the model hypothesis space:
\begin{equation}
\mathcal{L}{\text{topo}} = W_2 \big( \text{PD}(P_l), \text{PD}(P_s) \big)
\label{eq9}
\end{equation}
\(\text{PD}(\cdot)\) extracts persistence diagrams from predicted probability maps. \(W_2(\cdot,\cdot)\) is the second‑order Wasserstein distance measuring topological‑distribution discrepancy. For discrete persistence point‑sets \(X=\{x_i\}_{i=1}^m\), \(Y=\{y_j\}_{j=1}^n\):
\begin{equation}
W_2(X, Y) = \left( \min{\gamma \in \Pi(X,Y)} \sum_{i=1}^{m} \sum_{j=1}^{n} \gamma_{ij} |x_i - y_j|_2^2 \right)^{\frac{1}{2}}
\label{eq10}
\end{equation}
where \(\Pi(X,Y)\) denotes feasible coupling matrices and \(\gamma_{ij}\) stands for coupling weight entries.Bidirectional KL‑divergence knowledge distillation is applied to segmentation probability maps to mine implicit complementary information and align cross‑view semantics:
\begin{equation}
\begin{split}
\mathcal{L}_{\text{kd}} &= \frac{T^2}{2}\Big( \text{KL}\left(\text{softmax}(P_l/T) \parallel \text{softmax}(P_s/T)\right) \\
&\quad + \text{KL}\left(\text{softmax}(P_s/T) \parallel \text{softmax}(P_l/T)\right) \Big)
\end{split}
\label{eq11}
\end{equation}
where temperature T smooths soft‑label outputs while retaining fine structural details.

\section{Experiment}

\subsection{Datasets}

\begin{figure}[htbp] 
\vspace{-0.3cm}
\centering  
\centerline{\includegraphics[width=\linewidth]{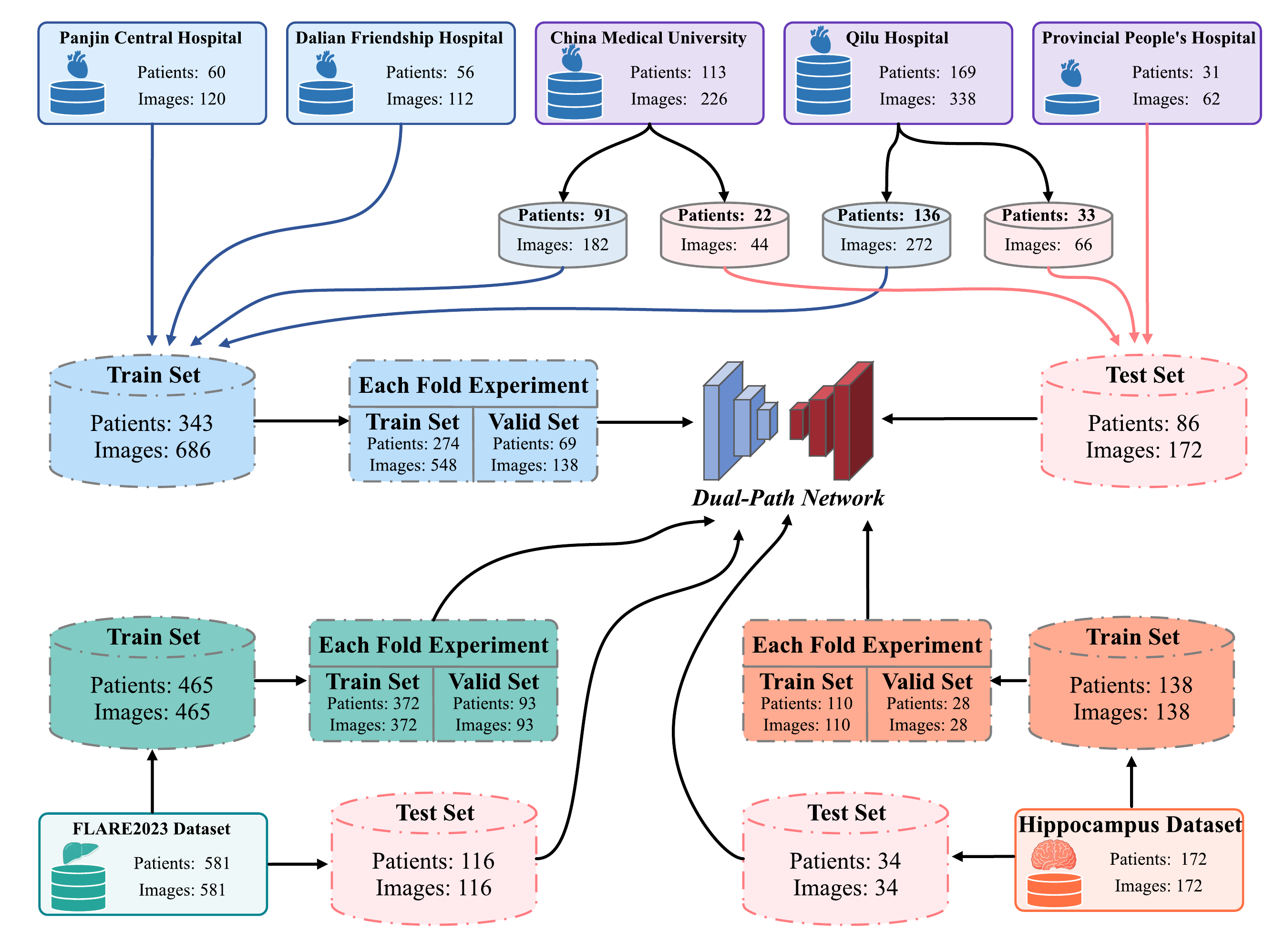}}
\caption{Details on the use of the three datasets. For each dataset, 20\% is randomly allocated as an independent test set, and the remaining data is used for training and validation using five-fold cross-validation.}
\vspace{-0.5cm}
\label{fig2}  
\end{figure}

We conducted our evaluation using three multi-view medical imaging datasets, including a private multicenter echocardiography dataset (referred to as PME), the publicly available FLARE2023 liver CT dataset, and the MSD hippocampal MRI dataset. The specific datasets used are shown in Fig \ref{fig2}.

PME Dataset: This dataset compiles data from 429 patients across five clinical centers, providing images of the apical four-chamber long-axis view and the papillary muscle-level short-axis view, along with annotations of the left ventricular myocardium. One image at the end of systole and one at the end of diastole are selected for each cardiac cycle, resulting in 858 images per view.

FLARE2023 Dataset\cite{15}: 2D slices with liver annotations from two views were extracted from the 3D abdominal CT volumes. After filtering based on liver size and the presence of tumors, 581 images were retained for each view for 581 patients.

MSD Dataset\cite{16}: We extracted 2D sections with dual views labeled for the anterior hippocampus. After image quality screening, 172 images were retained for each view for 172 patients.

To distinguish between the two views, the results are uniformly labeled as the “long-axis view” and the “short-axis view.” All datasets were split at the patient level to prevent data leaks. First, 20\% of the data was selected as an independent test set (not used for training), and the remaining data was subjected to five-fold cross-validation. All datasets underwent a standardized preprocessing workflow. For each dataset, images were cropped to a size of 512×512 pixels. During training, data augmentation was applied only to the training set to improve the model’s generalization ability. Neither the validation set nor the test set underwent data augmentation to ensure the objectivity and consistency of the evaluation.

\subsection{Evaluation Metrics}
We adopted four commonly used evaluation metrics in medical image segmentation: The Dice similarity coefficient(DSC), which measures the degree of overlap between the segmentation results and the gold standard. The Jaccard index (IoU), which is the ratio of the intersection to the union. The 95\% Hausdorff distance (HD95), which measures the degree of alignment of the segmentation boundaries. The average surface distance (ASD), which is the average distance between points on the segmentation boundaries. Using 5-fold cross-validation, we identified five optimal models, tested them on the independent test set, and reported the average values of the aforementioned metrics.

\subsection{Experimental Setup}
We implemented all networks using the PyTorch 1.12 framework. The Adam optimizer was used with an initial learning rate of 2e-4, which was reduced to a minimum learning rate of 1e-5 using a cosine annealing strategy. The batch size was set to 2, with a total of 100 training epochs. The backbone network was frozen for the first 30 epochs. The similarity threshold was set to \(\tau=0.2\), the distillation temperature to \(T=2.0\), and the loss weights to \(\alpha=0.1, \beta_s=0.05, \beta_l=0.05\). Except for specific parameters, the settings for the comparative experiments were identical to those of our method. All experiments were conducted on two NVIDIA RTX 4090 GPUs.

\begin{figure*}[htbp] 
\vspace{-0.3cm}
\centering  
\centerline{\includegraphics[width=\linewidth]{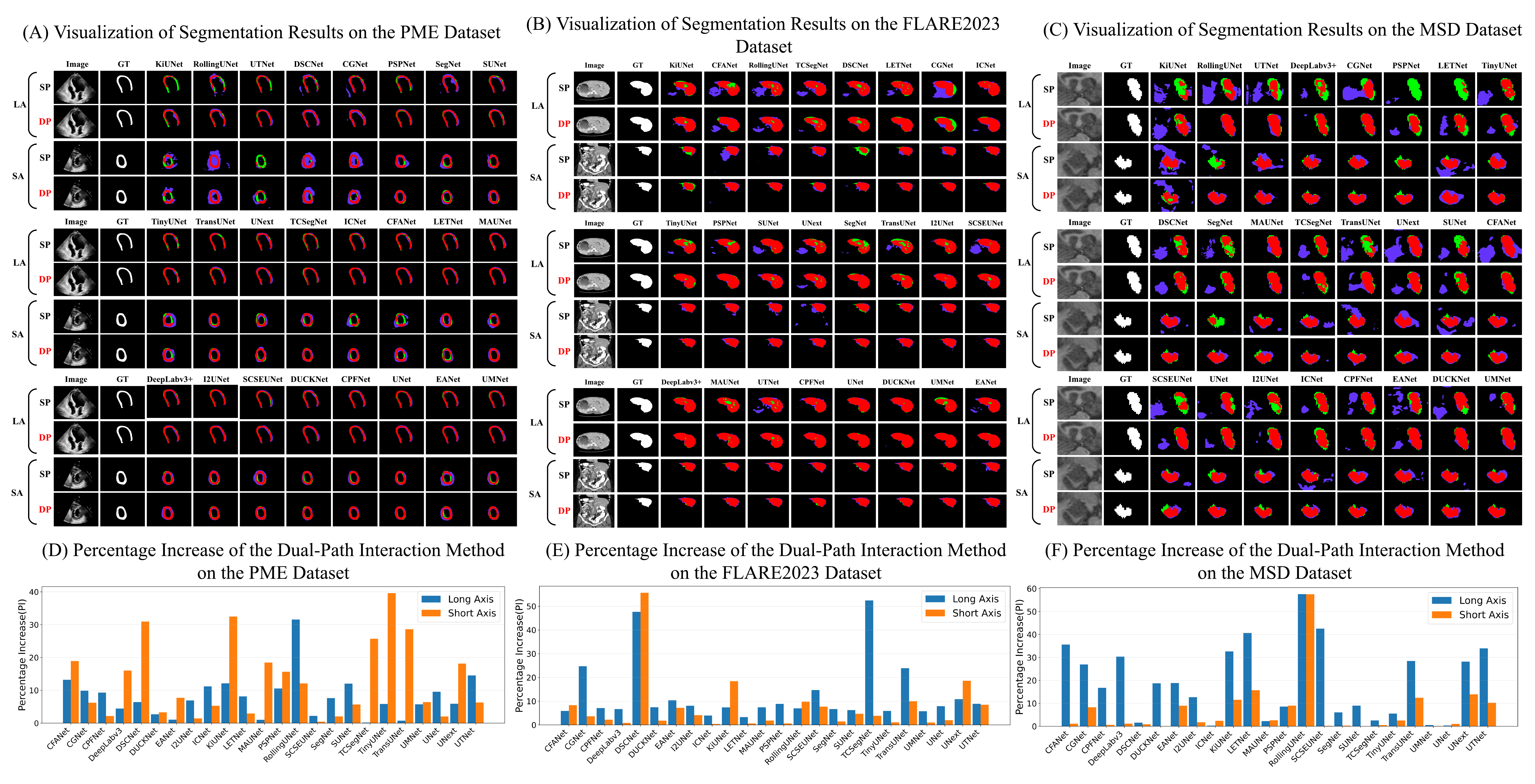}}
\caption{Visualization of the segmentation results and the bar chart showing the percentage increase in DSC. (A)–(C) show the validation results of the segmentation results against the mask, where red indicates the overlapping regions, purple indicates over-segmented regions, and green indicates under-segmented regions. (D)–(F) show the percentage improvements in the DSC metric achieved by the dual-path method.}
\vspace{-0.5cm}
\label{fig3}  
\end{figure*}

\begin{table*}[htbp]
\begin{center}
\caption{Results of the single-path and dual-path models on the PME dataset from the long-axis perspective. Here, PI represents the percentage improvement in the DSC metric achieved by the DP method relative to the SP method. The best result is shown in \colorbox[HTML]{FFCCC9}{red}, the second-best in \colorbox[HTML]{CBCEFB}{purple}, and the third-best in \colorbox[HTML]{96FFFB}{blue}.}\label{table1}  
\renewcommand{\arraystretch}{1.0}
\begin{adjustbox}{max width=\linewidth}
\begin{tabular}{c|ccccccccc|ccccccccc}
\hline
                                  & 
\multicolumn{9}{c|}{\textbf{Long Axis}}                                                                                                                                                                                                                                                                                                                & \multicolumn{9}{c}{\textbf{Short Axis}}                                                                                                                                                                                                                                                                                                                \\ \cline{2-19} 
                                  & 
\multicolumn{4}{c|}{\textbf{SP}}                                                                                                                        & \multicolumn{4}{c|}{\textbf{DP}}                                                                                                                        &                                    & \multicolumn{4}{c|}{\textbf{SP}}                                                                                                                        & \multicolumn{4}{c|}{\textbf{DP}}                                                                                                                        &                                    \\
\hline
\end{tabular}
\end{adjustbox}
\vspace{-0.8cm}
	\end{center}
\end{table*}

\section{Results}

\subsection{Comparison with Single-Path Segmentation Models}

To validate the DPIF’s generality and its advantages in few-shot scenarios, we selected 24 representative encoder-decoder networks as single-path (SP) baseline models and constructed corresponding dual-path (DP) variants by integrating the DPIF framework. All models were trained and evaluated on three multi-view datasets. The complete quantitative results are shown in TABLE \ref{table1} to TABLE \ref{table3} and Fig \ref{fig3}. Experimental results show that all dual-path variants achieved performance improvements on the three datasets. In terms of the DSC metric, the dual-path variants achieved an average improvement of 13.33\% on the long-axis views and 8.97\% on the short-axis views across the three datasets. Notably, the hippocampus long-axis dataset showed the largest average DSC improvement, reaching 19.2\%. Among these, RollingUNet demonstrated the most significant improvement, with a 57.57\% increase in DSC on the hippocampus long-axis dataset. This confirms that DPIF can effectively leverage the complementary anatomical information inherent in both views without introducing directional performance bias. Regarding boundary accuracy metrics, the average HD95 decreased by 4.27 in the long-axis view and by 2.38 in the short-axis view across the three datasets. It is worth noting that a few models (such as SCSEUNet) showed a slight decrease in HD95. Nevertheless, the overall trend confirms that DPIF effectively enhances the boundary continuity and anatomical fidelity of segmentation results for most architectures, delivering stable and consistent performance improvements across a variety of segmentation network architectures.

\begin{table*}[htbp]
\begin{center}
\caption{Results of the single-path and dual-path models on the FLARE2023 dataset.}\label{table2}  
\renewcommand{\arraystretch}{1.0}
\begin{adjustbox}{max width=\linewidth}
\begin{tabular}{c|ccccccccc|ccccccccc}
\hline
                                  & \multicolumn{9}{c|}{\textbf{Long Axis}}                                                                                                                                                                                                                                                                                                              & \multicolumn{9}{c}{\textbf{Short Axis}}                                                                                                                                                                                                                                                                                                              \\ \cline{2-19} 
                                  & \multicolumn{4}{c|}{\textbf{SP}}                                                                                                                       & \multicolumn{4}{c|}{\textbf{DP}}                                                                                                                       &                                    & \multicolumn{4}{c|}{\textbf{SP}}                                                                                                                       & \multicolumn{4}{c|}{\textbf{DP}}                                                                                                                       &                                    \\
\hline
\end{tabular}
\end{adjustbox}
\vspace{-0.8cm}
	\end{center}
\end{table*}

\begin{table*}[htbp]
\begin{center}
\caption{Results of the single-path and dual-path models on the MSD dataset.}\label{table3}  
\renewcommand{\arraystretch}{1.0}
\begin{adjustbox}{max width=\linewidth}
\begin{tabular}{c|ccccccccc|ccccccccc}
\hline
\multicolumn{1}{l|}{}                                  & \multicolumn{9}{c|}{\textbf{Long Axis}}                                                                                                                                                                                                                                                                                                              & \multicolumn{9}{c}{\textbf{Short Axis}}                                                                                                                                                                                                                                                                                                                 \\ \cline{2-19} 
\multicolumn{1}{l|}{}                                  & \multicolumn{4}{c|}{\textbf{SP}}                                                                                                                       & \multicolumn{4}{c|}{\textbf{DP}}                                                                                                                       &                                    & \multicolumn{4}{c|}{\textbf{SP}}                                                                                                                          & \multicolumn{4}{c|}{\textbf{DP}}                                                                                                                       &                                    \\
\hline
\end{tabular}
\end{adjustbox}
\vspace{-0.8cm}
	\end{center}
\end{table*}

\subsection{Comparison with Multi-View Segmentation Methods}

\begin{figure}[htbp] 
\vspace{-0.3cm}
\centering  
\centerline{\includegraphics[width=0.8\linewidth]{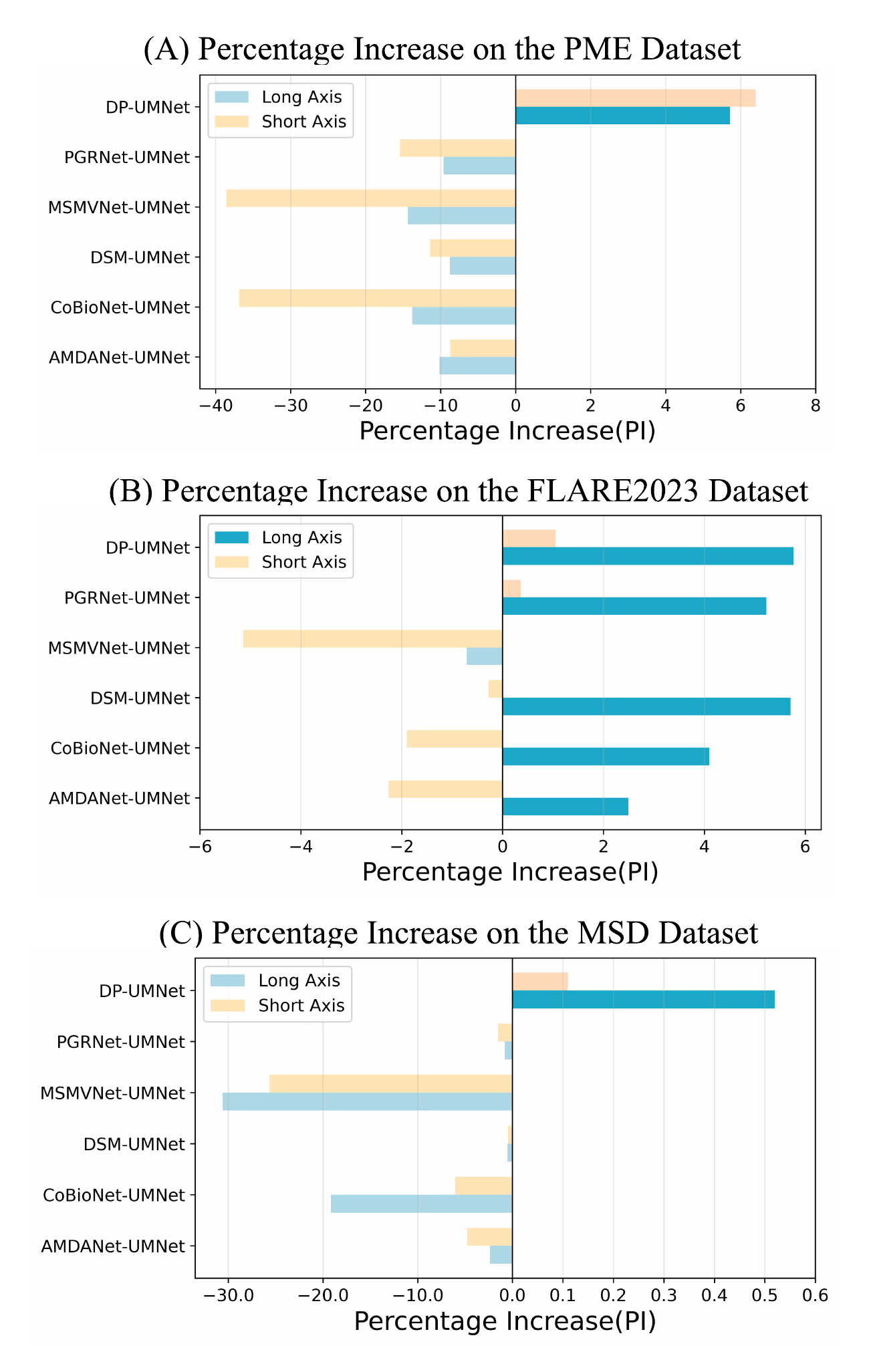}}
\caption{Comparison of Performance Improvements Achieved by the Multi-View Approach Across Three Datasets.}
\vspace{-0.5cm}
\label{fig4}  
\end{figure}

We selected five representative multi-view medical image segmentation methods for comparison. To ensure fairness in the experiments, all comparison methods used the same UMNet as the backbone network, were trained on exactly the same data split. The quantitative results are shown in TABLE \ref{table4}. Fig \ref{fig4} visualizes the percentage improvement for all methods. It can be seen that the DPIF framework demonstrates a consistent performance advantage across all views on the three datasets. In particular, on the long-axis view of the liver dataset, DP-UMNet achieves a 5.77\% improvement over the single-path baseline, significantly outperforming all comparison methods. In contrast, on all three test datasets, all other methods underperform DP-UMNet. Most of these methods even performed worse than the single-path UMNet baseline model, with only a few achieving a slight improvement on the liver dataset. Among them, MSMVNet and CoBioNet exhibited the most significant performance declines. Overall, compared to existing multi-view segmentation methods, the DPIF framework achieves stable and consistent performance improvements in multi-view tasks spanning different imaging modalities and anatomical structures.

\begin{table}[htbp]
\begin{center}
\caption{Comparison with Other Multi-View Segmentation Methods.}\label{table4}
  \renewcommand{\arraystretch}{1.2}
\begin{adjustbox}{max width=\columnwidth}
\begin{tabular}{c|cccccccccc}
\hline
                         & \multicolumn{10}{c}{LV dataset}                                                                                                                                \\ \cline{2-11} 
                         & \multicolumn{5}{c|}{Long Axis}                                                           & \multicolumn{5}{c}{Short Axis}                                      \\ \cline{2-11} 
\multirow{-3}{*}{Method} & DSC    & HD95    & ASD    & IOU     & \multicolumn{1}{c|}{PI(\%)}                        & DSC    & HD95    & ASD    & IOU    & PI(\%)                         \\ \hline
SP-UMNet                 & 0.8058 & 12.6179 & 3.2484 & 0.6748  & \multicolumn{1}{c|}{--}                             & 0.7984 & 12.5134 & 3.7446 & 0.6646 & --                              \\
AMDANet\cite{41}            & 0.7241 & 15.8557 & 4.9166 & 0.5680  & \multicolumn{1}{c|}{$-10.14$}                        & 0.7289 & 15.8725 & 4.9742 & 0.5734 & \cellcolor[HTML]{CBCEFB}$-8.70$   \\
CoBioNet\cite{42}           & 0.6948 & 16.2034 & 4.1474 & 0.5323  & \multicolumn{1}{c|}{$-13.78$}                        & 0.5041 & 20.4393 & 6.8256 & 0.3371 & $-36.86$                        \\
DSM\cite{43}                & 0.7351 & 15.7973 & 4.5805 & 0.5812  & \multicolumn{1}{c|}{\cellcolor[HTML]{CBCEFB}$-8.77$} & 0.7075 & 16.1543 & 4.9275 & 0.5474 & \cellcolor[HTML]{96FFFB}$-11.39$ \\
MSMVNet\cite{44}            & 0.6901 & 16.5968 & 5.3385 & 0.5267  & \multicolumn{1}{c|}{$-14.36$}                        & 0.4905 & 20.7283 & 7.1322 & 0.3249 & $-38.56$                         \\
PGRNet\cite{45}             & 0.7284 & 15.8456 & 4.9452 & 0.5728  & \multicolumn{1}{c|}{\cellcolor[HTML]{96FFFB}$-9.61$} & 0.6754 & 17.2599 & 5.5299 & 0.5097 & $-15.41$                         \\
DP-UMNet                 & 0.8519 & 11.1098  & 3.1702 & 0.7421  & \multicolumn{1}{c|}{\cellcolor[HTML]{FFCCC9}$+5.72$} & 0.8495 & 11.0153  & 3.1246 & 0.7384 & \cellcolor[HTML]{FFCCC9}$+6.40$   \\ \hline
                         & \multicolumn{10}{c}{FL dataset}                                                                                                                                \\ \cline{2-11} 
                         & \multicolumn{5}{c|}{Long Axis}                                                           & \multicolumn{5}{c}{Short Axis}                                      \\ \cline{2-11} 
\multirow{-3}{*}{Method} & DSC    & HD95    & ASD    & IOU     & \multicolumn{1}{c|}{PI(\%)}                        & DSC    & HD95    & ASD    & IOU    & PI(\%)                         \\ \hline
SP-UMNet                 & 0.9288 & 5.5999  & 2.0612 & 0.8672  & \multicolumn{1}{c|}{--}                             & 0.9404 & 4.4648  & 2.0467 & 0.8875 & --                              \\
AMDANet            & 0.9519 & 3.5387  & 1.8056 & 0.9098  & \multicolumn{1}{c|}{$+2.49$}                         & 0.9191 & 6.3722  & 2.1324 & 0.8504 & $-2.26$                          \\
CoBioNet           & 0.9669 & 3.1477  & 1.7196 & 0.9359  & \multicolumn{1}{c|}{$+4.10$}                          & 0.9225 & 5.6113  & 2.0414 & 0.8569 & $-1.90$                           \\
DSM                & 0.9818 & 2.3618  & 1.4945 & 0.9648  & \multicolumn{1}{c|}{\cellcolor[HTML]{CBCEFB}$+5.71$} & 0.9378 & 5.3893  & 1.9318 & 0.8826 & \cellcolor[HTML]{96FFFB}$-0.28$  \\
MSMVNet            & 0.9222 & 5.7246  & 2.0569 & 0.8562  & \multicolumn{1}{c|}{$-0.71$}                         & 0.8921 & 6.4842  & 2.4385 & 0.8053 & $-5.14$                          \\
PGRNet             & 0.9774 & 2.4743  & 1.5172 & 0.9555  & \multicolumn{1}{c|}{\cellcolor[HTML]{96FFFB}$+5.23$} & 0.9438 & 4.5985  & 2.0680 & 0.8935 & \cellcolor[HTML]{CBCEFB}$+0.36$  \\
DP-UMNet                 & 0.9824 & 2.2193  & 1.4525 & 0.9659  & \multicolumn{1}{c|}{\cellcolor[HTML]{FFCCC9}$+5.77$} & 0.9503 & 3.7747  & 1.6089 & 0.9060 & \cellcolor[HTML]{FFCCC9}$+1.05$  \\ \hline
                         & \multicolumn{10}{c}{Hip dataset}                                                                                                                               \\ \cline{2-11} 
                         & \multicolumn{5}{c|}{Long Axis}                                                           & \multicolumn{5}{c}{Short Axis}                                      \\ \cline{2-11} 
\multirow{-3}{*}{Method} & DSC    & HD95    & ASD    & IOU     & \multicolumn{1}{c|}{PI(\%)}                        & DSC    & HD95    & ASD    & IOU    & PI(\%)                         \\ \hline
SP-UMNet                 & 0.9077 & 6.4167  & 2.2079 & 0.8313  & \multicolumn{1}{c|}{--}                             & 0.9172 & 6.0197  & 2.1543 & 0.8477 & --                              \\
AMDANet            & 0.8863 & 6.9053  & 2.7587 & 0.7962  & \multicolumn{1}{c|}{$-2.36$}                         & 0.8733 & 7.1605  & 2.4956 & 0.7748 & $-4.79$                          \\
CoBioNet           & 0.7336 & 15.5838 & 4.3405 & 0.5795  & \multicolumn{1}{c|}{$-19.18$}                        & 0.8617 & 7.2136  & 2.5765 & 0.7573 & $-6.05$                          \\
DSM                & 0.9029 & 6.5943  & 2.4251 & 0.8232  & \multicolumn{1}{c|}{\cellcolor[HTML]{CBCEFB}$-0.53$} & 0.9131 & 6.2486  & 2.1561 & 0.8406 & \cellcolor[HTML]{CBCEFB}$-0.45$  \\
MSMVNet-UMNet            & 0.6300 & 17.6188 & 5.2609 & 0.4615  & \multicolumn{1}{c|}{$-30.59$}                        & 0.6819 & 16.0451 & 5.4792 & 0.5188 & $-25.65$                         \\
PGRNet             & 0.9004 & 6.6239  & 2.4178 & 0.82187 & \multicolumn{1}{c|}{\cellcolor[HTML]{96FFFB}$-0.80$}  & 0.9031 & 7.0377  & 2.3682 & 0.8235 & \cellcolor[HTML]{96FFFB}$-1.54$  \\
DP-UMNet                 & 0.9124 & 6.3704  & 2.1709 & 0.8395  & \multicolumn{1}{c|}{\cellcolor[HTML]{FFCCC9}$+0.52$} & 0.9182 & 1.9608  & 2.1767 & 0.8487 & \cellcolor[HTML]{FFCCC9}$+0.11$  \\ \hline
\end{tabular}
\end{adjustbox}
\vspace{-0.8cm}
	\end{center}
\end{table}

\begin{table*}[htbp]
\begin{center}
\caption{Ablation study of the main components in the DPIF framework. The first row indicates the use of only the single-path baseline model.}\label{table5}
  \renewcommand{\arraystretch}{1.0}
\begin{adjustbox}{max width=\linewidth}
\begin{tabular}{c|cccc|ccccc|ccccc}
\hline
\multicolumn{1}{l|}{}                         & \multicolumn{1}{l}{}                    & \multicolumn{1}{l}{}                    & \multicolumn{1}{l}{}                    & \multicolumn{1}{l|}{}                    & \multicolumn{5}{c|}{Long Axis}                                      & \multicolumn{5}{c}{Short Axis}                                      \\ \cline{6-15} 
\multicolumn{1}{l|}{\multirow{-2}{*}{Method}} & \multicolumn{1}{l}{\multirow{-2}{*}{A}} & \multicolumn{1}{l}{\multirow{-2}{*}{B}} & \multicolumn{1}{l}{\multirow{-2}{*}{C}} & \multicolumn{1}{l|}{\multirow{-2}{*}{D}} & DSC    & HD95    & ASD    & IOU    & PI(+\%)                             & DSC    & HD95    & ASD    & IOU    & PI(+\%)                             \\ \hline
                                              & $\times$                                       & $\times$                                       & $\times$                                       & $\times$                                        & 0.8058 & 12.6179 & 3.2484 & 0.6748 & --                              & 0.7984 & 12.5134 & 3.7446 & 0.6646 & --                              \\ \cline{2-15} 
                                              & $\checkmark$                                       & $\times$                                       & $\times$                                       & $\times$                                        & 0.8082 & 12.4836 & 3.2157 & 0.6785 & \cellcolor[HTML]{96FFFB}0.30 & 0.8011 & 12.3952 & 3.7024 & 0.6689 & \cellcolor[HTML]{96FFFB}0.34 \\
                                              & $\times$                                       & $\checkmark$                                       & $\times$                                       & $\times$                                        & 0.8245 & 10.7124 & 3.0842 & 0.7018 & \cellcolor[HTML]{FFCCC9}2.32 & 0.8192 & 10.5287 & 3.4519 & 0.6953 & \cellcolor[HTML]{FFCCC9}2.61 \\
                                              & $\times$                                       & $\times$                                       & $\checkmark$                                       & $\times$                                        & 0.8073 & 12.5471 & 3.2268 & 0.6772 & 0.19                         & 0.8003 & 12.4506 & 3.7185 & 0.6676 & 0.24                          \\
                                              & $\times$                                       & $\times$                                       & $\times$                                       & $\checkmark$                                        & 0.8152 & 11.2587 & 3.1395 & 0.6883 & \cellcolor[HTML]{CBCEFB}1.17 & 0.8097 & 11.1843 & 3.5426 & 0.6801 & \cellcolor[HTML]{CBCEFB}1.42 \\ \cline{2-15} 
                                              & $\times$                                       & $\checkmark$                                       & $\checkmark$                                       & $\checkmark$                                        & 0.8406 & 8.2875  & 2.8053 & 0.7251 & 4.32                         & 0.8372 & 8.3104  & 2.8479 & 0.7206 & 4.86                         \\
                                              & $\checkmark$                                       & $\times$                                       & $\checkmark$                                       & $\checkmark$                                        & 0.8337 & 9.0982  & 2.8704 & 0.7151 & 3.46                         & 0.8294 & 9.1257  & 2.9318 & 0.7097 & 3.88                         \\
                                              & $\checkmark$                                       & $\checkmark$                                       & $\times$                                       & $\checkmark$                                        & 0.8475 & 7.6391  & 2.7468 & 0.7350 & \cellcolor[HTML]{CBCEFB}5.18  & 0.8448 & 7.6823  & 2.7695 & 0.7312 & \cellcolor[HTML]{CBCEFB}5.81 \\
                                              & $\checkmark$                                       & $\checkmark$                                       & $\checkmark$                                       & $\times$                                        & 0.8452 & 8.3957  & 2.8164 & 0.7317 & \cellcolor[HTML]{96FFFB}4.89 & 0.8419 & 8.4236  & 2.8351 & 0.7275 & \cellcolor[HTML]{96FFFB}5.45 \\
\multirow{-10}{*}{UMNet}                      & $\checkmark$                                       & $\checkmark$                                       & $\checkmark$                                       & $\checkmark$                                        & 0.8519 & 7.2098  & 2.7102 & 0.7415 & \cellcolor[HTML]{FFCCC9}5.72  & 0.8495 & 7.2153  & 2.7246 & 0.7379 & \cellcolor[HTML]{FFCCC9}6.40 \\ \hline
\end{tabular}
\end{adjustbox}
\vspace{-0.8cm}
	\end{center}
\end{table*}

\subsection{Ablation Studies and Analyses}

To quantitatively verify individual contributions and synergies among DPIF components, we perform progressive ablation experiments on the multicenter PME echocardiography dataset with UMNet as the backbone. Taking single‑path UMNet as baseline, nine ablation variants are built to evaluate four modules: two‑layer feature decoupling (A), Sinkhorn unpaired distribution alignment (B), adaptive interaction weights (C), and cross‑view topological consistency regularization (D). Quantitative results are reported in TABLE \ref{table5}. The results show that decoupling features across two layers yields only a slight improvement, as it achieves feature separation and noise suppression within branches but does not facilitate cross-view information exchange. Sinkhorn-based distribution alignment is the primary driver of performance improvement, increasing DSC by 2\% for the long-axis view and 2.24\% for the short-axis view, respectively. This validates that distribution-level semantic matching is the primary channel for cross-view knowledge transfer. When used alone, adaptive interaction weights yielded a negligible improvement, indicating that dynamic weighting cannot function effectively without the support of distribution alignment. Cross-view topological regularization independently optimized segmentation boundaries, increasing the long-axis DSC by 1.14\% and reducing HD95 by approximately 0.39.

\subsection{Analysis of Different Training Data Usage}

\begin{figure}[htbp] 
\vspace{-0.3cm}
\centering  
\centerline{\includegraphics[width=\linewidth]{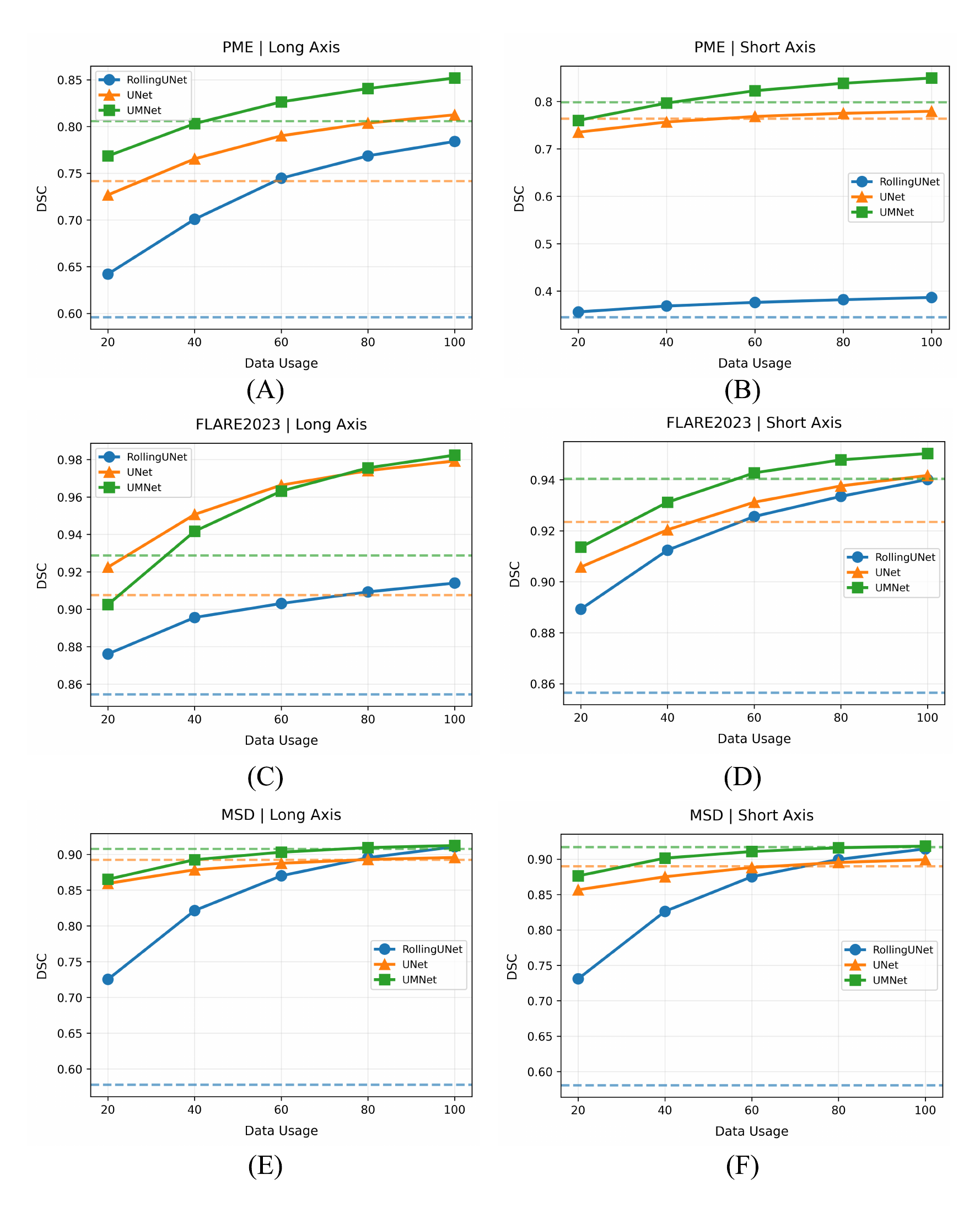}}
\caption{Ablation analysis regarding the amount of training data. The dashed line represents the model’s DSC for training on the entire dataset in a single-path pipeline.}
\vspace{-0.5cm}
\label{fig5}  
\end{figure}

To investigate the performance and few-shot adaptability of DPIF under different levels of data usage, we conducted gradient data experiments on three multi-view datasets using three representative backbone networks (RollingUNet, UNet and UMNet). Among these, RollingUNet represents a class of networks with poor baseline performance, UNet represents a class of classic segmentation networks, and UMNet represents a class of networks with superior baseline performance.
We employed a stratified inclusive sampling method, stratified by multicenter origin and lesion proportion, to construct training subsets of 20\%, 40\%, 60\%, and 80\%. Smaller subsets were fully nested within larger ones to isolate data volume as the sole variable while preserving the statistical distribution of the full dataset. All experiments used a fixed test set and five-fold cross-validation. The corresponding performance curves are illustrated in Fig \ref{fig5}. The results show that the performance of the dual-path model increases monotonically as the amount of training data increases. The performance improvement is most significant in the range from 20\% to 40\%. In most cases, the DPIF-enhanced model can achieve the same accuracy as the full-data single-path model even when the annotation rate is below 60
For weaker backbone networks, DPIF delivers greater performance improvements. DP-RollingUNet, trained on only 20\% of the labeled data, outperformed the full-data single-path baseline model on the PME short-axis view. Cross-modal analysis indicates that the benefits of few-shot learning are positively correlated with task difficulty. In the PME echocardiography task, DP-UMNet outperforms full-data single-path models with a 60\% annotation rate. On the high-contrast FLARE2023 liver CT dataset, it requires only 40\% annotated data. On a high-resolution hippocampal MRI dataset, DP-UMNet requires a 60\% annotation rate to approach the performance of the full-data single-path model.

\subsection{Blind Review of Segmentation Results by Clinicians}

To verify whether the improvements achieved through dual-pathway interaction could translate into clinically usable segmentation results, we conducted a blinded review study across three datasets. Three ultrasound specialists and three radiology specialists evaluated the segmentation results using a 1–5-point Likert scale for 30 randomly selected cases from each view. For each case, the original image and the segmentation results from both DP-UMNet and SP-UMNet were displayed simultaneously, with the positions of the results randomized to avoid subjective bias. The results are shown in Table\ref{table6}. The advantages of DP-UMNet were most pronounced in the sections with the weakest baseline performance: in the PME short-axis section, DP-UMNet raised the average score to 3.49 ± 0.60, and the proportion of cases with a score of $\ge$4 increased from 24.4\% to 52.2\%. In cases where single-path baseline models had already achieved high clinical scores (e.g., FLARE2023 short-axis and MSD short-axis), the dual-path enhancement did not yield further significant improvements. The inter-reader consistency (ICC) across views ranged from moderate to high. These results confirm that the introduction of DPIF not only improves the agreement between segmentation results and ground-truth labels but also demonstrates that DPIF achieves consistent clinical validation.

\begin{table*}[htbp]
\begin{center}
\caption{Comparison of subjective scores and inter-rater reliability for DP and SP across datasets and views.}\label{table6}
  \renewcommand{\arraystretch}{1.0}
\begin{adjustbox}{max width=\linewidth}
\begin{tabular}{c|c|ccccccc}
\hline
Dataset                & View  & Mean±SD(DP) & Mean±SD(SP) & Wilcoxon p (n=30) & Cliff's $\delta$ & Mixed-effects $\beta$ (p)     & DP-ICC(2,3) & SP-ICC(2,3) \\ \hline
\multirow{2}{*}{PME}   & Long  & 3.90±0.85   & 3.79±0.97   & 0.647             & 0.17      & 0.11 (0.341)            & 0.857       & 0.830       \\
                       & Short & 3.49±0.60   & 3.07±0.67   & 0.004             & 0.57      & 0.42 (\textless{}0.001) & 0.767       & 0.687       \\ \hline
\multirow{2}{*}{FLARE} & Long  & 4.40±0.82   & 4.00±0.89   & 0.035             & 0.40      & 0.40 (\textless{}0.001) & 0.837       & 0.717       \\
                       & Short & 4.23±0.85   & 4.21±0.79   & 0.917             & 0.10      & 0.02 (0.819)            & 0.815       & 0.813       \\ \hline
\multirow{2}{*}{MSD}   & Long  & 4.23±0.81   & 3.96±0.70   & 0.049             & 0.47      & 0.28 (0.001)            & 0.811       & 0.806       \\
                       & Short & 4.21±0.74   & 4.14±0.77   & 0.519             & 0.00      & 0.07 (0.465)            & 0.674       & 0.632       \\ \hline
\end{tabular}
\end{adjustbox}
\vspace{-0.8cm}
	\end{center}
\end{table*}

\section{Discussion}
In the comparative experiment, traditional multi-view methods cannot guarantee consistent performance improvements and sometimes even underperform compared to single-view baseline models, particularly on noisier, unpaired data. Thanks to its multiple functional modules—particularly feature alignment and two-layer decoupling, DPIF mitigates the effects of noise and differences in interaction targets. As a result, it continues to demonstrate performance improvements even on challenging datasets such as ultrasound and MRI.

Gradient data experiments confirm that DPIF improves data utilization efficiency. The magnitude of performance improvement increases as task difficulty and baseline model performance decrease. For low-performance backbone networks such as RollingUNet, DPIF delivers significant improvements, indicating its immense potential in the field of rare diseases where annotations are scarce. Notably, performance continues to improve steadily even when the data usage is 100\%. Therefore, DPIF serves not only as a solution for few-shot learning but also as a general-purpose enhancement module for fully supervised multi-view clinical pipelines.

Nevertheless, the framework presented in this paper has some limitations. The current implementation supports only two-view interactions, extending it to three or more views would facilitate the volumetric segmentation of 3D data. Topological regularization employs a differentiable persistent homology approximation, which may overlook subtle topological features. Therefore, a more precise differentiable topological solver is needed.

\section{Conclusion}

This paper proposes a general-purpose, plug-and-play dual-path interaction framework called DPIF for multi-view medical image segmentation, which improved data utilization when data is limited. DPIF achieves consistent performance improvements across 24 encoder-decoder backbone networks and outperforms existing multi-view segmentation methods. It can match or exceed the performance of single-path models trained on the full dataset using only a subset of data of a certain size. This demonstrates that DPIF provides a practical and scalable solution for multi-view medical image segmentation under data-scarce conditions.

\bibliographystyle{plainnat}
\bibliography{ref}

\end{document}